\documentclass{llncs}

\usepackage{makeidx}

\usepackage[utf8x]{inputenc}

\usepackage[T1, T2A]{fontenc}

\usepackage{graphicx}

\usepackage{hyperref}

\usepackage{fancyvrb,newverbs,xcolor}
\definecolor{cverbbg}{gray}{0.93}

\begin{document}

\pagestyle{headings}

\title{Towards Self-explanatory Ontology Visualization with Contextual Verbalization
\thanks{This work has been supported by the ESF project 2013/0005/1DP/1.1.1.2.0/13/APIA/VIAA/049 and the Latvian State Research program NexIT project No.1 ``Technologies of ontologies, semantic web and security''.}}

\author{Renārs Liepiņš \and Uldis Bojārs \and Normunds Grūzītis \and Kārlis Čerāns \and Edgars Celms}
\institute{Institute of Mathematics and Computer Science, University of Latvia
\email{renars.liepins@lumii.lv}, \email{uldis.bojars@lumii.lv},  \email{normunds.gruzitis@lumii.lv}, \email{karlis.cerans@lumii.lv}, \email{edgars.celms@lumii.lv}
}

\maketitle

\begin{abstract}

Ontologies are one of the core foundations of the Semantic Web. To participate in Semantic Web projects, domain experts need to be able to understand the ontologies involved. Visual notations can provide an overview of the ontology and help users to understand the connections among entities. However, the users first need to learn the visual notation before they can interpret it correctly. Controlled natural language representation would be readable right away and might be preferred in case of complex axioms, however, the structure of the ontology would remain less apparent.
We propose to combine ontology visualizations with contextual ontology verbalizations of selected ontology (diagram) elements, displaying controlled natural language (CNL) explanations of OWL axioms corresponding to the selected visual notation elements. Thus, the domain experts will benefit from both the high-level overview provided by the graphical notation and the detailed textual explanations of particular elements in the diagram.

\end{abstract}

\section{Introduction}

Semantic Web technologies have been successfully applied in pilot projects and are now transitioning toward mainstream adoption in the industry. However, for this transition to go successfully, there are still hurdles that have to be overcome. One of them are the difficulties that domain experts have in understanding mathematical formalisms and their notations that are used in ontology engineering.

Visual notations have been proposed as a way to help domain experts to work with ontologies. Indeed, when domain experts collaborate with ontology experts in designing an ontology ``they very quickly move to sketching 2D images to communicate their thoughts'' \cite{howse2011visualizing}. The use of diagrams has also been supported by an empirical study done by Warren et al. where they reported that ``one-third of [participants] commented on the value of drawing a diagram'' to understand what is going on in the ontology \cite{warren2014usability}.

Despite the apparent success of the graphical approaches, there is still a fundamental problem with them. When a novice user wants to understand a particular ontology, he or she cannot just look at the diagram and know what it means. The user first needs to learn the syntax and semantics of the notation -- its mapping to the underlying formalism. In some diagrams an edge with a label \emph{P} between nodes \emph{A} and \emph{B} might denote a property \emph{P} that has domain \emph{A} and range \emph{B}, while in others it might mean that every \emph{A} has at least one property \emph{P} to something that is \emph{B}. This limitation has long been noticed in software engineering \cite{siau2004informational} and, for this reason, formal models in software engineering are often translated into informal textual documentation by systems analysts, so that they can be validated by domain experts \cite{frederiks2006information}.

A similar idea of automatic conversion of ontologies into seemingly informal controlled natural language (CNL) texts and presenting the texts to domain experts has been investigated by multiple groups \cite{power2010expressing,stevens2011automating,kuhn2013understandability}. CNL is more understandable to domain experts and end-users than the alternative representations because the notation itself does not have to be learned, or the learning time is very short. Hoverer, the comparative studies of textual and graphical notations have shown that while domain experts that are new to graphical notations better understand the natural language text, they still prefer the graphical notations in the long run \cite{ottensooser2012making,sharafi2013empirical}. It leads to a dilemma of how to introduce domain experts to ontologies. The CNL representation shall be readable right away and might be preferred in case of complex axioms (restrictions) while the graphical notation makes the overall structure and the connections more comprehensible.

We present an approach that combines the benefits of both graphical notations and CNL verbalizations. The solution is to extend the graphical notation with contextual verbalizations of the axioms that are represented by the selected graphical element. The graphical representation gives the users an overview of the ontology while the contextual verbalizations explain what the particular graphical elements mean. Thus, domain experts that are novices in ontology engineering shall be able to learn and use the graphical notation rapidly and independently without special training.

In Section~\ref{sect:Proposal}, we present the general principles of extending graphical ontology notations with contextual natural language verbalizations. In Section~\ref{sect:CaseStudy}, we demonstrate the proposed approach in practice by extending a particular graphical ontology notation and editor, OWLGrEd, with contextual verbalizations in controlled English. In Section~\ref{sect:Discussion}, we discuss the benefits and limitations of our approach, as well as sketch some future work. Related work is discussed in Section~\ref{sect:RelatedWork}, and we conclude the article in Section~\ref{sect:Conclusions}.

\section{Extending Graphical Notations with Contextual Verbalizations}
\label{sect:Proposal}

This section describes the proposed approach for contextual verbalization of graphical elements in ontology diagrams, starting with a motivating example. We are focusing particularly on OWL ontologies, assuming that they are already given and that the ontology symbols (names) are lexically motivated and consistent, i.e., we are not considering the authoring of ontologies in this article, although the contextual verbalizations might be helpful in the authoring process as well, and it would motivate to follow a lexical and consistent naming convention.

\subsection{Motivating Example}

\begin{figure}
    \centering
    \includegraphics[width=1.0\textwidth]{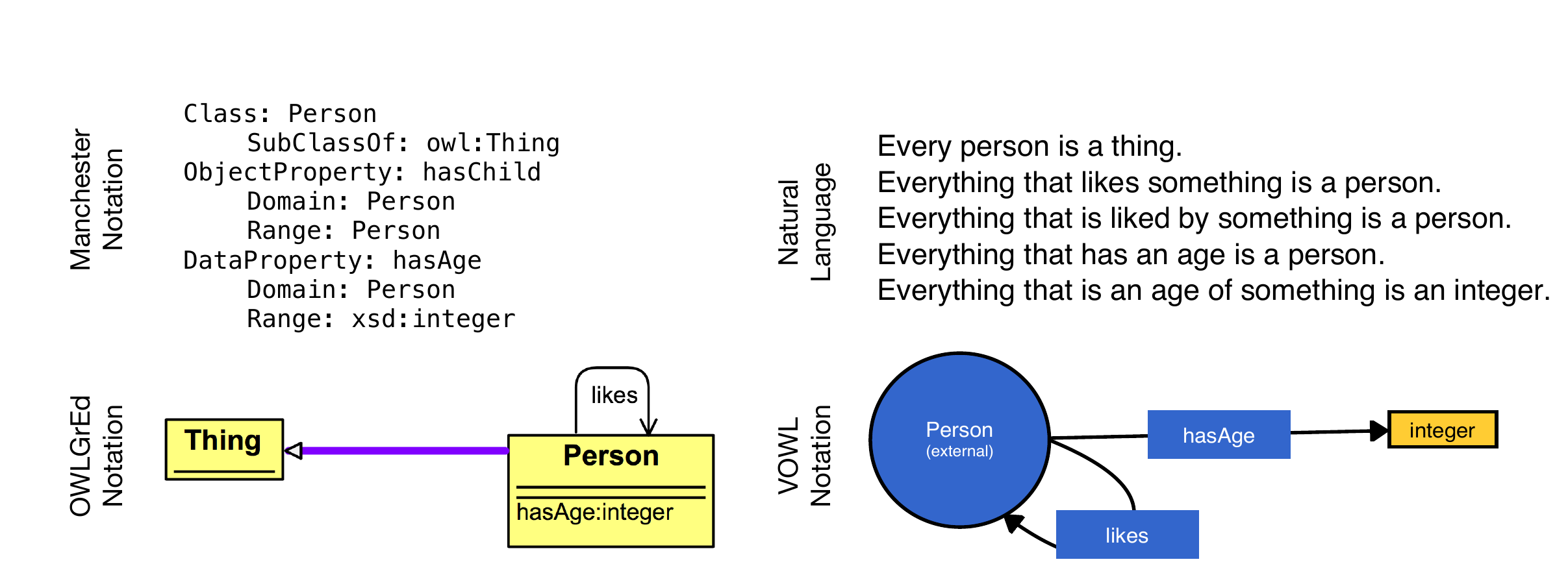}
    \caption{A simplified ontology fragment alternatively represented by using Manchester OWL Syntax, VOWL and OWLGrEd, and an explanation in a controlled natural language}
    \label{fig:BasicOntologyExample}
\end{figure}

In most diagrammatic OWL ontology notations, object property declarations are shown either as boxes (for example in VOWL \cite{lohmann2014vowl}) or as labeled links connecting the property domain and range classes as in OWLGrEd \cite{barzdicnvs2010uml}. Figure~\ref{fig:BasicOntologyExample} illustrates a simplified ontology fragment that includes classes \emph{Person} and \emph{Thing}, an object property \emph{likes} and a data property \emph{hasAge}. This fragment is represented by using three alternative formal notations: Manchester OWL Syntax \cite{horridge2009owl}, VOWL and OWLGrEd. As can be seen, the visualizations are tiny and may already seem self-explanatory. Nevertheless, even in this simple case, the notation for domain experts may be far from obvious. For example, the Manchester OWL Syntax uses the terms \emph{domain} and \emph{range} when defining a property, and these terms may not be familiar to a domain expert. In the graphical notations, the situation is even worse because the user may not even suspect that the edges represent more than one assertion and that the assertions are far-reaching. In the case of \emph{likes}, it means that everyone that likes something is \emph{necessarily} a person, and vice versa.

We have encountered such problems in practice when introducing ontologies in the OWLGrEd notation to users familiar with the UML notation. Initially, it turned out that they are misunderstanding the meaning of the association edges. For example, they would interpret that the edge \emph{likes} in Figure~\ref{fig:BasicOntologyExample} means ``persons \emph{may} like persons'', which is true, however, they would also assume that other disjoint classes could also have this property, which is false in OWL because multiple domain/range axioms of the same property are combined to form an intersection. Thus, even having a very simple ontology, there is a potential for misunderstanding the meaning of both the formal textual notation (e.g., Manchester OWL Syntax) and the graphical notations.

The data property \emph{hasAge} in Figure~\ref{fig:BasicOntologyExample} illustrates another kind of a problem. In some graphical notations (e.g., VOWL), data properties are represented by edges, and their value types -- by nodes (using a style that is different from class nodes). In other notations (e.g., OWLGrEd), data properties are represented by labels inside the class node that corresponds to the data property's domain. While the representation and therefore the reading is similar to the object properties in VOWL, verbalization might help to novice OWLGrEd users.

\subsection{Proposed Approach}

We propose to extend graphical ontology diagrams with contextual on-demand verbalizations of OWL axioms related to the selected diagram elements, with the goal to help users to better understand their ontologies and to learn the graphical notations based on their own and/or real-world examples.

The contextual verbalization of ontology diagrams relies on the assumption that every diagram element represents a set of ontology axioms, i.e., the ontology axioms are generally presented locally in the diagram, although possibly a single ontology axiom can be related to several elements of the diagram.

The same verbalization can be applied to all the different OWL visual notations, i.e., we do not have to design a new verbalization (explanation) grammar for each new visual notation, because they all are mapped to the same underlying OWL axioms. Thus, the OWL visualizers can reuse the same OWL verbalizers to provide contextual explanations of any graphical OWL notation.

\begin{figure}
    \centering
    \includegraphics[width=0.47\textwidth]{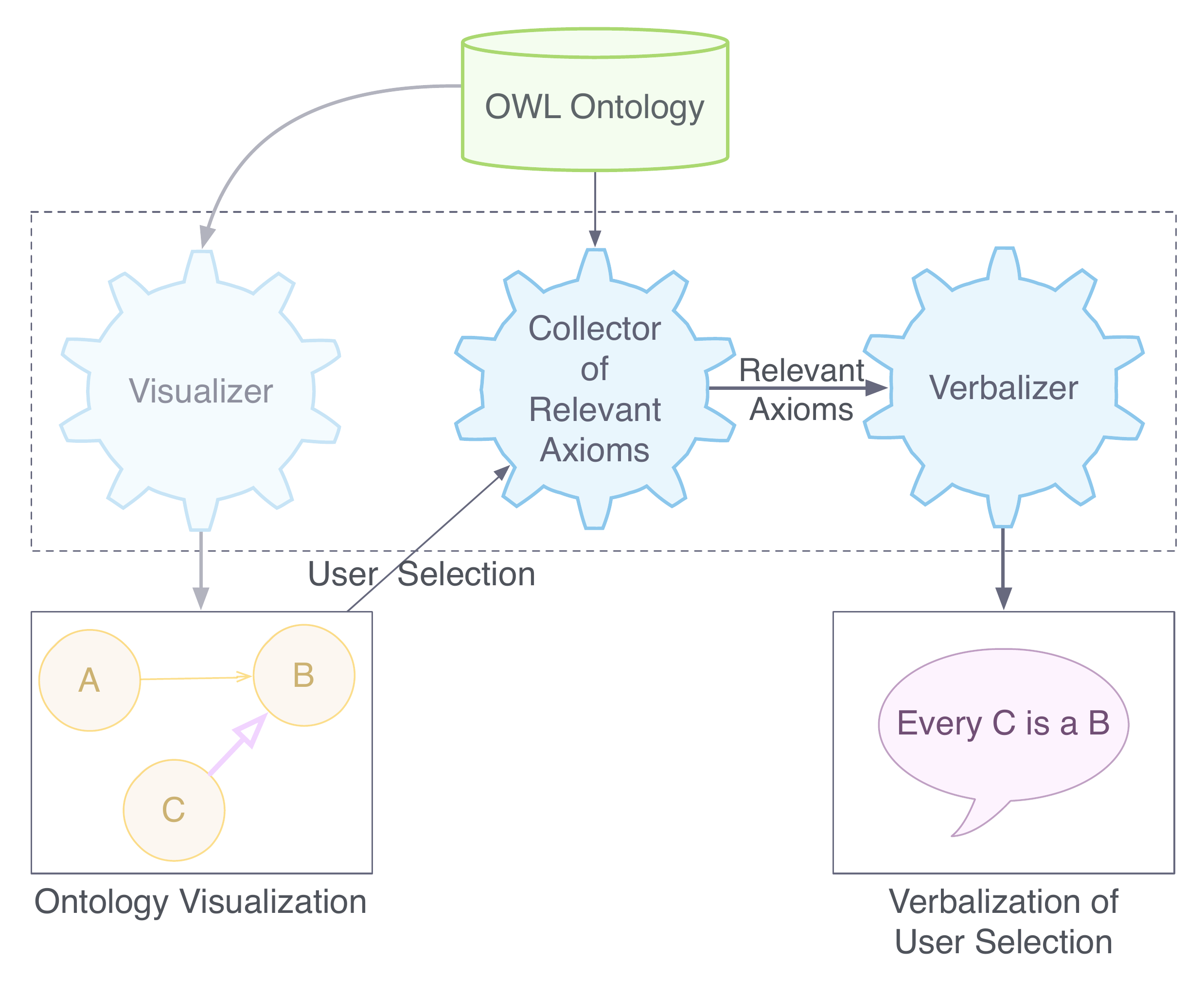}
    \caption{Architecture of a contextual ontology verbalizer}
    \label{fig:Architecture}
\end{figure}

By reusing ontology verbalizers, existing ontology visualization systems can be easily extended with a verbalization service. Figure~\ref{fig:Architecture} illustrates the proposed approach:

\begin{enumerate}
  \item \emph{Visualizer} is the existing visualization component that transforms an OWL ontology into its graphical representation.
  \item The system is extended by a \emph{User Selection} mechanism that allows users to select the graphical element that they want to verbalize.
  \item \emph{Collector} gathers a subset of the ontology axioms that correspond to the selected graphical element.
  \item The relevant axioms are passed to \emph{Verbalizer} that produces CNL statements -- a textual explanation that is shown to the user.
\end{enumerate}

By applying the proposed approach and by using natural language to interactively explain what the graphical notation means, developers of graphical OWL editors and viewers can enable users (domain experts in particular) to avoid misinterpretations of ontology elements and their underlying axioms, resulting in a better understand of both the ontology and the notation.
For example, when domain experts encounter the ontology in Figure~\ref{fig:BasicOntologyExample}, they would not have to guess what the elements of this graphical notation mean. Instead, they can just ask the system to explain the notation using the example of the ontology that they are exploring. When the user clicks on the edge \emph{likes} in Figure~\ref{fig:BasicOntologyExample}, the system shows the verbalization
\begin{quote}
\emph{Everything that likes something is a person. Everything that is liked by something is a person.}
\end{quote}
which unambiguously explains the complete meaning of this graphical element.

The verbalization of ontology axioms has been shown to be helpful in teaching OWL to newcomers both in practical experience reports \cite{rector2004owl} as well as in statistical evaluations \cite{kuhn2013understandability}.

\section{Case Study: Extending OWLGrEd with Contextual Verbalizations in ACE}
\label{sect:CaseStudy}

The proposed approach is illustrated by a case study demonstrating the enhancement of extending OWLGrEd, a graphical notation and editor for OWL, with on-demand contextual verbalizations of the underlying OWL axioms using Attempto Controlled English (ACE)~\cite{fuchs2008attempto}. The CNL verbalization layer allows users to inspect a particular element of the presented ontology diagram and to receive a verbal explanation of the ontology axioms that are related to this ontology element.

A demonstration of our implementation is available online.\footnote{\url{http://owlgred.lumii.lv/cnl-demo}}


\subsection{Overview of the OWLGrEd Notation}

The OWLGrEd notation \cite{barzdicnvs2010uml} is a compact and complete UML-style notation for OWL 2 ontologies. It relies on Manchester OWL Syntax \cite{horridge2009owl} for certain class expressions. 

This notation is implemented in the OWLGrEd ontology editor\footnote{\url{http://owlgred.lumii.lv}} and its online ontology visualization tool\footnote{\url{http://owlgred.lumii.lv/online_visualization/}} \cite{liepins2015owlgred}. The approach proposed in this article is implemented as a custom version of the OWLGrEd editor and visualization tool.

In order to keep the visualizations compact and to give the ontology developers flexibility in visualizing their ontologies, the OWLGrEd notation provides several alternatives how a certain OWL axiom can be represented (e.g., either as a visual element or as an expression in  Manchester OWL Syntax inside the class element). This makes the OWLGrEd notation a good case study for exploring the use of contextual verbalizations. In order to fully understand a visualization of an ontology, the domain experts would need to understand both the graphical elements and the expressions in Manchester OWL Syntax.

\begin{figure*}[t]
    \centering
    \includegraphics[width=\textwidth]{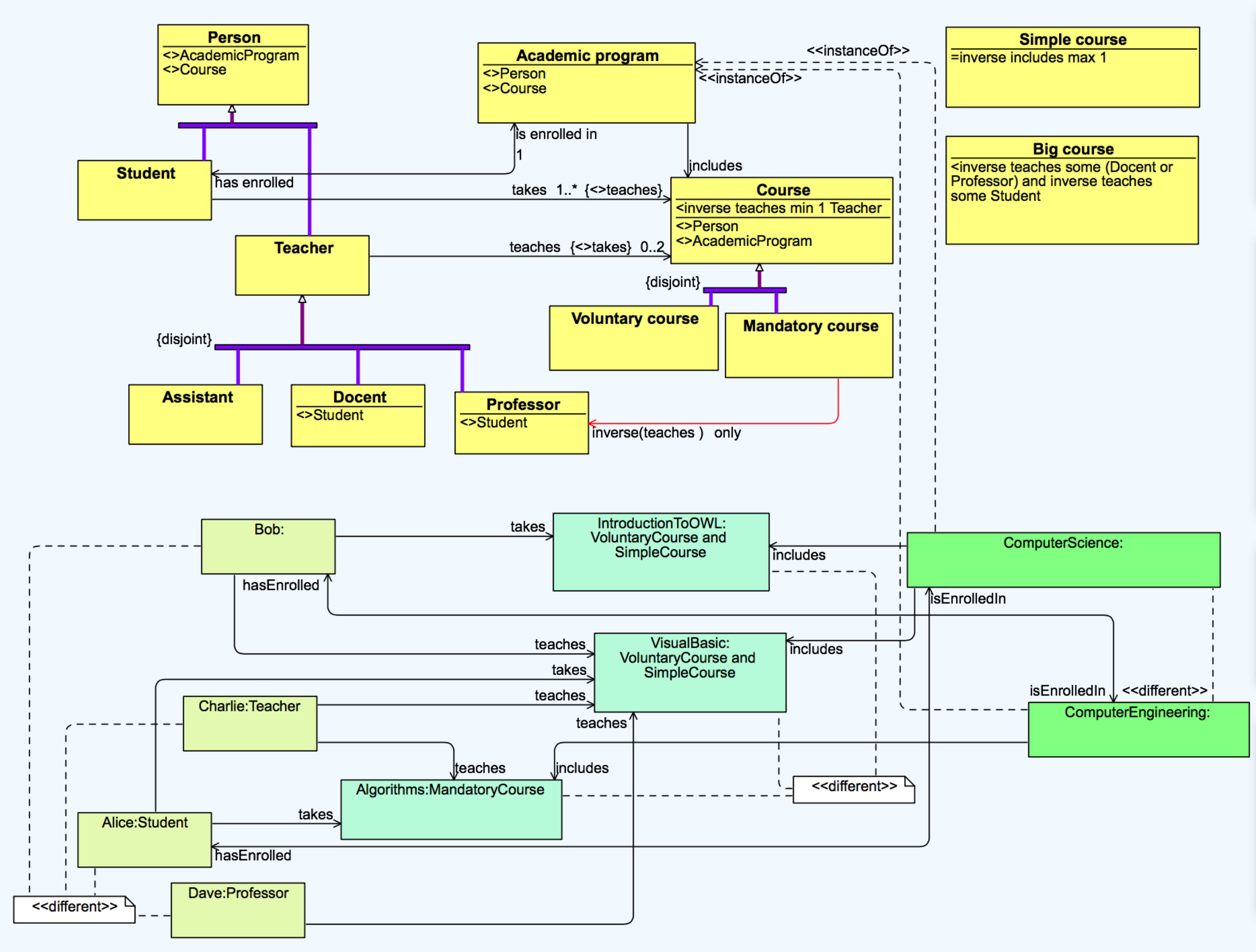}
    \caption{An example mini-university ontology represented in the OWLGrEd notation}
    \label{fig:OWLGrEdNotationExample}
\end{figure*}

Figure~\ref{fig:OWLGrEdNotationExample} demonstrates the OWLGrEd notation through an example of a mini-university ontology. OWL classes (e.g., \emph{Student}, \emph{Person}, \emph{Course} in Figure \ref{fig:OWLGrEdNotationExample}) are represented by UML classes while OWL object properties are represented by roles on the associations between the relevant domain and range classes (e.g., \emph{teaches}, \emph{takes}, \emph{hasEnrolled}). OWL datatype properties are represented by attributes of the property domain classes. OWL individuals are represented by UML objects (e.g., \emph{Alice}, \emph{Bob}, \emph{ComputerScience}). 

Simple cardinality constraints can be described along with the object or datatype properties (e.g., \emph{every student is enrolled in exactly 1 academic program}), and \emph{inverse-of} relations can be encoded as inverse roles of the same association.

Subclass assertions can be visualized in the form of UML generalizations that can be grouped together using generalization sets (the ``fork'' elements). Disjointness or completeness assertions on subclasses can be represented using UML generalization set constraints (e.g., classes \emph{Assistant}, \emph{Docent} and \emph{Professor} all are subclasses of \emph{Teacher} and are pairwise disjoint). OWLGrEd also introduces a graphical notation for property-based class restrictions -- a red arrow between the nodes. For example, the red arrow between classes \emph{MandatoryCourse} and \emph{Professor} corresponds to the following restriction in the Manchester notation:
\begin{quote}
\texttt{MandatoryCourse SubClassOf inverse teaches only Professor}
\end{quote}

Class elements can have text fields with OWL class expressions in Manchester OWL Syntax. While OWLGrEd allows to specify class expressions in a graphical form, more compact visualizations can be achieved by using the textual Manchester notation (e.g., in the descriptions of \emph{Course}, \emph{SimpleCourse} and \emph{BigCourse}). The ‘\textless’ textual notation is used for sub-class and sub-property relations, ‘=’ for equivalent classes/properties and ‘\textless\textgreater’ for disjoint classes/properties (e.g., \emph{Person} \textless\textgreater \emph{Course}).

\begin{figure*}[t]
    \centering
    \includegraphics[width=\textwidth]{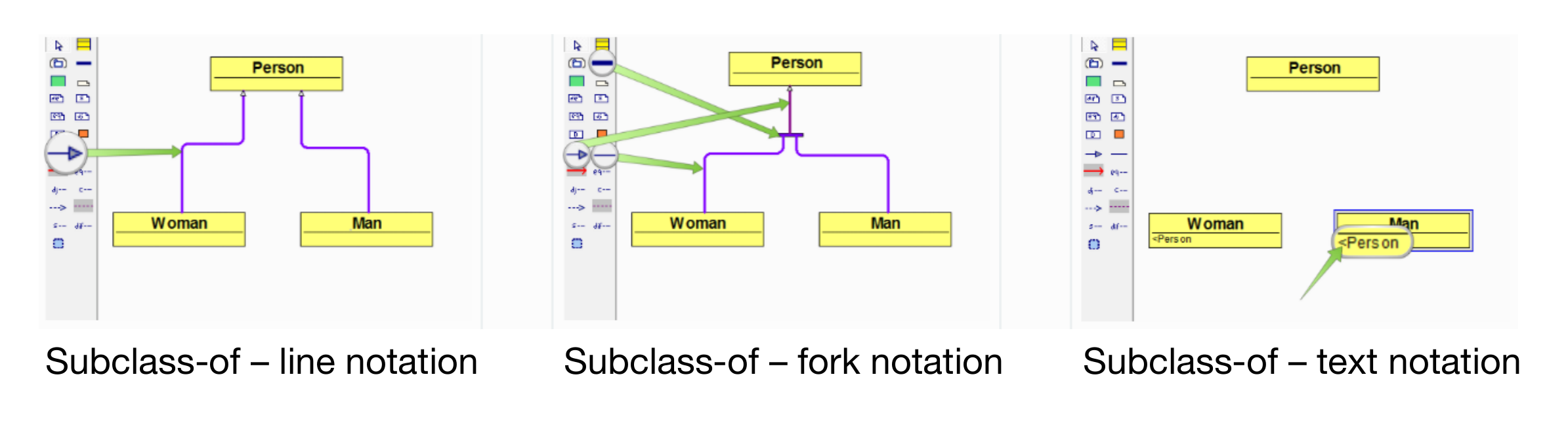}
    \caption{Options for representing the class hierarchy in OWLGrEd}
    \label{fig:MultipleWaysToShowSubclassRel}
\end{figure*}

Figure~\ref{fig:MultipleWaysToShowSubclassRel} illustrates the multiple ways how OWL axioms can be represented in diagrams using the OWLGrEd notation. It shows how the generalization (a \emph{subclass-of} relation) can be represented using a line notation, a more compact ``fork'' notation and a text notation that may be preferable in some cases (e.g., for referring to a superclass that is defined using an OWL class expression and is not referenced anywhere else).

On the assertion or individual level (ABox), there are two options for stating class assertions (instances): by using the \emph{instanceOf} arrow to the corresponding class element or by stating the class name or expression in the box element denoting the individual.

We refer to \cite{barzdicnvs2010uml,barzdicnvs2010owlgred} for a more detailed explanation of the OWLGrEd notation and editor, as well as the principles of its visual extensions \cite{cerans2012advanced}.

\subsection{Adding Verbalizations to OWLGrEd}

In order to help domain experts to understand the visualized ontology diagram, they are presented with explanations -- textual representation (verbalization) of all OWL axioms corresponding to a given element of the ontology diagram.

The verbalization can help users even in relatively simple cases, such as object property declarations where user's intuitive understanding of the domain and range of the property might not match what is asserted in the ontology. The verbalization of OWL axioms makes this information explicit while not requiring users to be ontology experts. The value of contextual ontology verbalization is even more apparent for elements whose semantics might be somewhat tricky even for more experienced users (e.g., \emph{some}, \emph{only} and cardinality constraints on properties, or generalization forks with \emph{disjoint} and \emph{complete} constraints).

The CNL verbalization layer for which an experimental support has been added to the OWLGrEd editor enhances the ontology diagrams with an interactive means for viewing textual explanations of the axioms associated with a particular graphical element.

By clicking a mouse pointer on an element, a pop-up widget is thrown, containing a CNL verbalization of the corresponding axioms in  Attempto Controlled English. By default, the OWLGrEd visualizer minimizes the number of verbalization widgets shown simultaneously by hiding them after a certain timeout. For users to simultaneously see the verbalizations for multiple graphical elements, there is an option to ``freeze'' the widgets and prevent them from disappearing.

\begin{figure*}[t]
    \centering
    \includegraphics[width=0.95\textwidth]{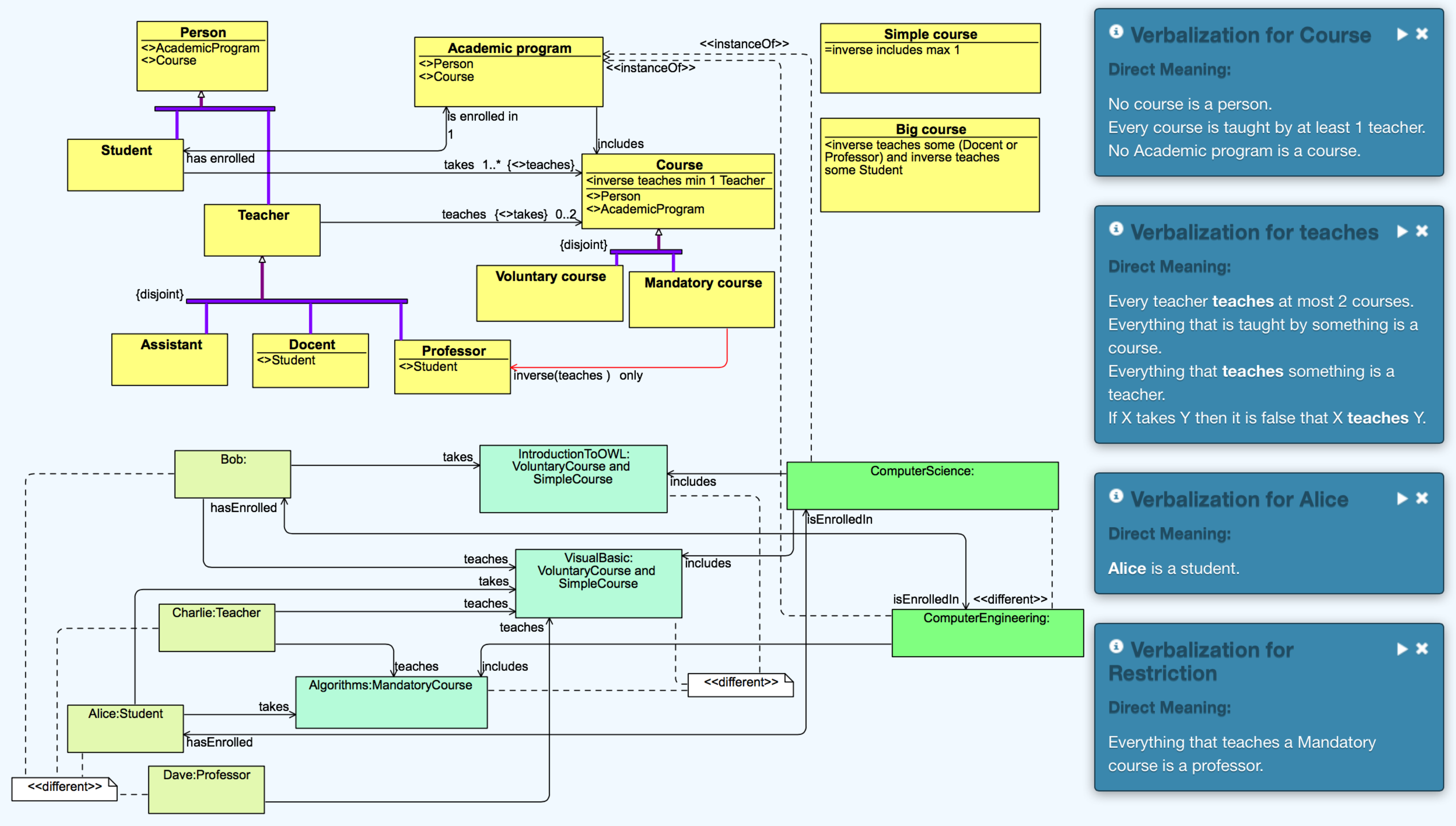}
    \caption{The example ontology in the OWLGrEd notation (see Figure~\ref{fig:OWLGrEdNotationExample}) with CNL verbalizations (explanations) of the selected diagram elements.}
    \label{fig:CNLExampleWithMultiplePopups}
\end{figure*}

Figure~\ref{fig:CNLExampleWithMultiplePopups} shows an example of multiple verbalizations displayed on the diagram introduced in Figure~\ref{fig:OWLGrEdNotationExample}. They describe the ontology elements that represent the class \emph{Course}, the object property \emph{teaches}, the individual \emph{Alice} and the restriction on the class \emph{MandatoryCourse}. Verbalizations are implicitly linked to the corresponding elements using the element labels. While it might be less convenient to identify the implicit links in a static image, the interactive nature of the combined ontology visualization and verbalization tool makes it easier for users to keep the track. Visual cues (e.g., a line between a verbalization and the diagram element) could be added to make the linking more noticeable. However, to keep the visualization simple, such cues are not currently employed.

The object property \emph{teaches}, represented in the diagram by an edge connecting the class \emph{Teacher} to the class \emph{Course}, has the following verbalization in ACE (see Figure~\ref{fig:CNLExampleWithMultiplePopups}):

\begin{samepage}

\begin{quote}
\emph{Every teacher teaches at most 2 courses.}
\end{quote}
\begin{quote}
\emph{Everything that is taught by something is a course.}
\end{quote}
\begin{quote}
\emph{Everything that teaches something is a teacher.}
\end{quote}
\begin{quote}
\emph{If X takes Y then it is false that X teaches Y.}
\end{quote}

\end{samepage}

Note that the specific OWL terms, like \emph{disjoint}, \emph{subclass} and \emph{inverse}, are not used in the ACE statements. The same meaning is expressed implicitly~-- via paraphrasing~-- using more general common sense constructions and terms.

In this case, the edge represents not only the domain and range axioms of the property but also the cardinality of the range and the restriction that \emph{teaches} is disjoint with \emph{takes} (expressed by the if-then statement).

The property restriction on the class \emph{MandatoryCourse}, shown in the diagram as a red line connecting the class \emph{MandatoryCourse} to the class \emph{Professor}, is another case when a CNL explanation is essential. Its meaning is expressed in ACE by the following statement:
\begin{quote}
\emph{Everything that teaches a mandatory course is a professor.}
\end{quote}

In this case, similarly as in the case of the \emph{disjoint} restrictions, the ACE verbalizer has rewritten the axiom in a more general but semantically equivalent form avoiding the use of the determiner \emph{only} (\emph{nothing but} in ACE)~\cite{kaljurand:phd}. At the first glance, it might seem confusing for an expert, however, such a semantic paraphrase can be helpful to better understand the consequences of the direct reading of the axiom:
\begin{quote}
\emph{Every mandatory course is taught by nothing but professors.}
\end{quote}

The steps involved in the verbalization of an ontology diagram element (as implemented in OWLGrEd) are:

\begin{enumerate}
  \item Every diagram element represents a set of OWL axioms.
  \item The set of axioms corresponding to this element is ordered by axiom type and sent as a list to the verbalization component.
  \item The verbalization component returns a corresponding list of CNL statements.
  \item The resulting CNL statements (the textual explanation of the diagram element) are displayed to the user.
\end{enumerate}

The translation from OWL to ACE is done by reusing the readily available verbalizer from the ACE toolkit~\cite{kaljurand2007verbalizing}.\footnote{\url{http://attempto.ifi.uzh.ch/site/resources/}}

In order to acquire lexically and grammatically well-formed sentences (from the natural language user's point of view), additional lexical information may need to be provided, e.g., that the property \emph{teaches} is verbalized using the past participle form ``taught'' in the passive voice (\emph{inverse-of}) constructions or that the class \emph{MandatoryCourse} is verbalized as a multi-word unit ``mandatory course''. This information is passed to the OWL-to-ACE verbalizer as an ontology-specific lexicon.

In the case of controlled English, the necessary lexical information can be largely restored automatically from the entity names (ontology symbols), provided that English is used as a meta-language and that the entity names are lexically motivated and consistently formed.

If aiming for multilingual verbalizations, domain-specific translation equivalents would have to be specified additionally, which, in general, would be a semi-automatic task.

An appropriate and convenient means for implementing a multilingual OWL verbalization grammar is Grammatical Framework (GF)~\cite{Ranta2004} which provides a reusable resource grammar library for about 30 languages.\footnote{\url{http://www.grammaticalframework.org/}} Moreover, an ACE grammar library based on the GF general-purpose resource grammar library is already available for about 10 languages~\cite{CamilleriEtAl2012}. This allows for using English-based entity names and the OWL subset of ACE as an interlingua, following the two-level OWL-to-CNL approach suggested in~\cite{GruzitisBarzdins2011}.

In fact, we have applied the GF-based approach to provide an optional support for lexicalization and verbalization in OWLGrEd in both English and Latvian, a highly inflected Baltic language.

\section{Discussion}
\label{sect:Discussion}

This section discusses the use of contextual ontology verbalization, focusing on its applicability to various graphical notations, extending the scope of axioms to include in verbalizations and the potential limitations of the approach.

\subsection{Applicability to Other Notations}
\label{sec-other-notations}

The proposed approach is applicable to any ontology visualization where graphical elements represent one or more OWL axioms. The value of using verbalization functionality is higher for more complex notations (e.g., OWLGrEd) where graphical elements may represent multiple axioms but even in simple graphical notations, where each graphical element corresponds to one axiom, users will need to know how to read the notation. Contextual ontology verbalization addresses this need by providing textual explanations of diagram elements and the underlying OWL axioms.

A more challenging case is notations where some OWL axioms are represented as spatial relations between the elements and are not directly represented by any graphical elements (e.g., Concept Diagrams represent \emph{subclass-of} relations as shapes that are included in one another~\cite{stapleton2014vision}). In order to represent these axioms in ontology verbalization they need to be ``attached'' to one or more graphical elements that these axioms refer to. As a result, they will be included in verbalizations of relevant graphical elements. In the case of Concept Diagrams, the \emph{subclass-of} relation, which is represented by shape inclusion, would be verbalized as part of the subclass shape.

\subsection{Extending the Scope of Verbalization}

The scope of OWL axioms that are included in CNL explanations of diagram elements can be adjusted by modifying the \emph{Collector} component (see Figure~\ref{fig:Architecture}) that selects OWL axioms related to a particular element. In our primary use case, the choice of OWL axioms to verbalize is straightforward – for each element only the axioms directly associated with this element (i.e. the axioms that this element represents) are used for generating CNL verbalizations. Depending on the graphical notation the scope of axioms for verbalization may need to be expanded, as we pointed out in Section~\ref{sec-other-notations}.

Enlarging the scope of axioms to verbalize may also be useful for generating contextual documentation of a selected element. In contrast to the primary use case, where the verbalization was used to explain the notation, in the case of contextual documentation, we want to show the user all the axioms that are related to the selected element (e.g., when the user selects a class node, the system would also show verbalizations of domain and range axioms where this class is mentioned). Such approach is widely used in textual ontology documentation (e.g., in FOAF vocabulary specification\footnote{\url{http://xmlns.com/foaf/spec/\#sec-crossref}}).

Another use case for enhanced verbalizations is running inference on the ontology and including inferred OWL axioms in verbalizations. Since it is not practical to show all inferred axioms in the graphical representation, contextual verbalizations are a suitable place for displaying this information. Verbalization of inferred axioms may also be useful for ontology debugging. For instance, in the ontology in Figure \ref{fig:OWLGrEdNotationExample}, it might be derived that every individual belonging to the class \emph{BigCourse} also belongs  to the class \emph{Course}, however the same would not necessarily be true for every individual belonging to the class \emph{SimpleCourse} (one can term this a design error present in the ontology).

To support these use cases, the scope of verbalization axioms can be increased to include additional axioms. It is important to note that by doing this (i.e. by adding axioms not directly represented by the element) we would lose the benefit of having equivalent visual and verbal representations of the ontology and the resulting verbalizations might not be as useful for users in learning the graphical notation. This limitation can be partially alleviated by visually distinguishing between CNL sentences that represent direct axioms (i.e. axioms that directly correspond to a graphical element) and other axioms included in the verbalization.

\subsection{Limitations}

Due to the interactive nature of the approach it might not work well for documenting ontologies when diagrams are printed out or saved as screenshots. While the example in Figure \ref{fig:CNLExampleWithMultiplePopups} shows how multiple verbalizations are displayed simultaneously and that they can still be useful when saved as screenshots, the image would become too cluttered if a larger amount of verbalizations were displayed simultaneously therefore a naive approach of showing all verbalizations on a diagram at once would not work for documentation.

The combined ontology visualization and verbalization approach can be adap\-ted to documenting ontologies by exporting fragments of the ontology diagram showing a particular graphical element along with verbalized statements corresponding to this element. The resulting documentation would have some redundancy because one CNL statement may be relevant to multiple concepts. However, it has been shown that such ``dictionaries'' are perceived to be more usable than the alternative, where all axioms verbalizations are displayed just once, without grouping \cite{williams2011levels}.

Verbalization techniques that are a part of the proposed approach have the same limitations as ontology verbalization in general. In particular, verbalization may require additional lexical information to generate grammatically well-formed sentences. To some degree, by employing good ontology design practices and naming conventions as well as by annotating ontology entities with lexical labels, this limitation can be overcome.
Another issue is specific kinds of axioms that are difficult or verbose to express in natural language without using terms of the underlying formalism.

As it was mentioned, the contextual verbalizations could be generated in multiple languages, provided that the translation equivalents have been provided while authoring the ontology. This would allow domain experts to explore ontologies in their native language and would be even more important to the regular end-users exploring ontology-based applications.

\section{Related Work}
\label{sect:RelatedWork}

To the best of our knowledge, there are no publications proposing combining OWL ontology visualizations with contextual CNL verbalizations but there has been a movement towards cooperation between both fields. In ontology visualizations, notations have been adding explicit labels to each graphical element that describes what kind of axiom it represents. For example, in VOWL a line representing a \emph{subclass-of} relation is explicitly labeled with the text ``Subclass of''. This practice makes the notation more understandable to users as reported in the VOWL user study where a user stated that ``there was no need to use the printed [notation reference] table as the VOWL visualization was very self-explanatory'' \cite{lohmann2014vowl}. However, such labeling of graphical elements is only useful in notations where each graphical element represents one axiom. In more complex visualizations where one graphical element represents multiple axioms there would be no place for all the labels corresponding to these axioms. For example, in the OWLGrEd notation, class boxes can represent not just class definitions but also \emph{subclass-of} and \emph{disjoint classes} assertions. In such cases, verbalizations provide understandable explanations. Moreover, in some notations (e.g., Concept Diagrams~\cite{stapleton2014vision}) there might be no graphical elements at all for certain kinds of axioms, as it was mentioned in Section \ref{sec-other-notations}.

In the field of textual ontology verbalizations there has been some exploration of how to make verbalizations more convenient for users. One approach that has been tried is grouping verbalizations by entities. It produces a kind of a dictionary, where records are entities (class, property, individual), and every record contains verbalizations of axioms that refer to this entity. The resulting document is significantly larger than a plain, non-grouped verbalization because many axioms may refer to multiple entities and thus will be repeated in each entity. Nevertheless, the grouped presentation was preferred by users \cite{williams2011levels}. Our approach can be considered a generalization of this approach, where a dictionary is replaced by an ontology visualization that serves as a map of the ontology.

An ad-hoc combination of verbalization and visualization approaches could be achieved using existing ontology tools such as  Prot\'eg\'e by using separate visualization and verbalization plugins (e.g., ProtégéVOWL\footnote{\url{http://vowl.visualdataweb.org/protegevowl.html}} for visualization and ACEView\footnote{\url{http://attempto.ifi.uzh.ch/aceview/}} for verbalization). However, this would not help in understanding the graphical notation because the two views are independent, and thus a user cannot know which verbalizations correspond to which graphical elements. Our approach employs closer integration of the two ontology representations and provides contextual verbalization of axioms that directly correspond to the selected graphical element, helping users in understanding the ontology and learning the graphical notation used.

\section{Conclusions}
\label{sect:Conclusions}

Mathematical formalisms used in ontology engineering are hard to understand for domain experts. Usually, graphical notations are suggested as a solution to this problem. However, the graphical notations, while preferred by domain experts, still have to be learned to be genuinely helpful in understanding. Until now the only way to learn these notations was by reading the documentation.

In this article, we proposed to use the CNL verbalizations to solve the learning problem. Using our approach the domain expert can interactively select a graphical element and receive the explanation of what the element means. The explanation is generated by passing the corresponding axioms of the element through one of the existing verbalization services. The service returns natural language sentences explaining the OWL axioms that correspond to the selected element and thus explaining what it means.

We demonstrated the proposed approach in a case study where we extended an existing ontology editor with the contextual CNL explanations. We also presented the architecture of the extension that is general enough to apply to wide range of other ontology notations and tools.

In conclusion, we have shown how to extend graphical notations with contextual CNL verbalizations that explain the selected ontology element. The explanations help domain experts to rapidly and independently learn and use the notation from the beginning without a special training, thus making it easier for domain experts to participate in ontology engineering without extended training, which solves one of the problems that hinder the adoption of Semantic Web technologies in the mainstream industry.

\bibliographystyle{splncs03}
\bibliography{bibliography}

\end{document}